\pdfoutput=1

\documentclass[11pt]{article}

\usepackage{ACL2023}  

\usepackage{times}
\usepackage{latexsym}

\usepackage[T1]{fontenc}

\usepackage[utf8]{inputenc}

\usepackage{microtype}

\usepackage{inconsolata}

\usepackage{multirow}
\usepackage{tabularx}
\usepackage{textcomp}
\usepackage{listings}
\usepackage{booktabs}

\usepackage{makecell}
\usepackage[ruled,linesnumbered]{algorithm2e}
\usepackage{hyperref}
\usepackage{graphicx}
\newcolumntype{Y}{>{\centering\arraybackslash}X}
\usepackage{enumerate}
\usepackage{amsmath}
\newcommand{\state}[1]{\textit{``#1''}}
\usepackage{xspace}
\newcommand{\schema}{\texttt{Schema.org}\xspace}
\newcommand{\doid}{\texttt{DOID}\xspace}
\newcommand{\foodon}{\texttt{FoodOn}\xspace}
\newcommand{\go}{\texttt{GO}\xspace}

\newcommand{\inter}[1]{#1^\mathcal{I}}

\usepackage{titlesec}
\titleformat{\subsubsection}[runin]{\normalfont\normalsize\bfseries}{\thesubsubsection}{1em}{}[]

\titlespacing{\subsubsection}{0pt}{1.0ex plus 1.0ex minus .2ex}{1.5ex plus .2ex}

\usepackage{paralist}

\usepackage{amsthm}
\newtheorem*{definition*}{Definition (Assumed Disjointness)}

\definecolor{pptgreen}{RGB}{112, 173, 71}
\definecolor{pptred}{RGB}{192, 0, 0}
\definecolor{pptblue}{RGB}{68, 114, 196}

\usepackage{float}

\title{Language Model Analysis for Ontology Subsumption Inference}

\author{Yuan He\textsuperscript{\rm 1},
 Jiaoyan Chen\textsuperscript{\rm 2}, 
 Ernesto Jim\'{e}nez-Ruiz\textsuperscript{\rm 3,4}, \\
 {\bf Hang Dong\textsuperscript{\rm 1},}
 {\bf Ian Horrocks\textsuperscript{\rm 1}} \\
 \textsuperscript{\rm 1} University of Oxford, \textsuperscript{\rm 2} The University of Manchester, \\
 \textsuperscript{\rm 3} City, University of London, \textsuperscript{\rm 4} University of Oslo \\
 \texttt{\{yuan.he,hang.dong,ian.horrocks\}@cs.ox.ac.uk}\\
 \texttt{jiaoyan.chen@manchester.ac.uk}\\
 \texttt{ernesto.jimenez-ruiz@city.ac.uk}
 }

\begin{document}
\maketitle
\begin{abstract}
Investigating whether pre-trained language models (LMs) can function as knowledge bases (KBs) has raised wide research interests recently. However, existing works focus on simple, triple-based, relational KBs, but omit more sophisticated, logic-based, conceptualised KBs such as OWL ontologies. 
To investigate an LM's knowledge of ontologies, we propose \textsc{OntoLAMA}, a set of inference-based probing tasks and datasets from ontology subsumption axioms involving both atomic and complex concepts\footnote{An ontology \textit{concept} is also known as a \textit{class}. To avoid confusion with \textit{class} in machine learning classification, we stick to use the term \textit{concept}.}. 
We conduct extensive experiments on ontologies of different domains and scales, and our results demonstrate that LMs encode relatively less background knowledge of Subsumption Inference (SI) than traditional Natural Language Inference (NLI) but can improve on SI significantly when a small number of samples are given. 
We will open-source our code and datasets.\footnote{Code and Instructions: \url{https://krr-oxford.github.io/DeepOnto/ontolama}; Dataset at HuggingFace: \url{https://huggingface.co/datasets/krr-oxford/OntoLAMA/} or at Zenodo: \url{https://doi.org/10.5281/zenodo.6480540}}
\end{abstract}

\section{Introduction}


The advancements of large pre-trained language models (LMs) have sparked research interests in investigating how much explicit semantics LMs can learn or infer from knowledge bases (KBs) \cite{AlKhamissi2022ARO}. The LAMA (LAnguage Model Analysis) probe \cite{petroni-etal-2019-language} is among the first works that adopt prompt-based methods to simulate the process of querying relational knowledge from various KBs such as ConceptNet \cite{conceptnet} and GoogleRE\footnote{\url{https://code.google.com/archive/p/relation-extraction-corpus/}}. Some subsequent studies focus on probing specific types of knowledge from sources like commonsense KBs \cite{Da2021AnalyzingCE}, biomedical KBs \cite{sung-etal-2021-language}, temporal KBs \cite{dhingra-etal-2022-time}, and cross-lingual KBs \cite{Liu2021EnhancingML}. 

However, existing ``LMs-as-KBs'' works focus on simple, triple-based, relational KBs, but neglect more formalised, logic-based, conceptualised KBs. For example, a statement like \state{London is the capital of the UK} can be expressed in the triple \texttt{(London, capitalOf, UK)}; but a sentence like \state{arthritis is a kind
of arthropathy with an inflammatory morphology}, which describes the concept \state{arthritis}, cannot be easily expressed using just triples. Conceptual knowledge like this requires a formal and expressive representation to be defined precisely. A well-known model for conceptual knowledge is the OWL\footnote{For simplicity, we refer to the second edition OWL~2 as OWL: \url{https://www.w3.org/TR/owl2-overview/}} ontology \cite{bechhofer2004owl,owl2}, which can be seen as a description logic (DL) KB with rich built-in vocabularies for knowledge representation and various reasoning tools supported. Taking the example of \state{arthritis}, in DL the concept can be described as $Arthritis \sqsubseteq Arthropathy \sqcap \exists hasMorphology.Inflammatory$.

\begin{figure}[t!]
\begin{center}
    \includegraphics[width=0.49\textwidth]{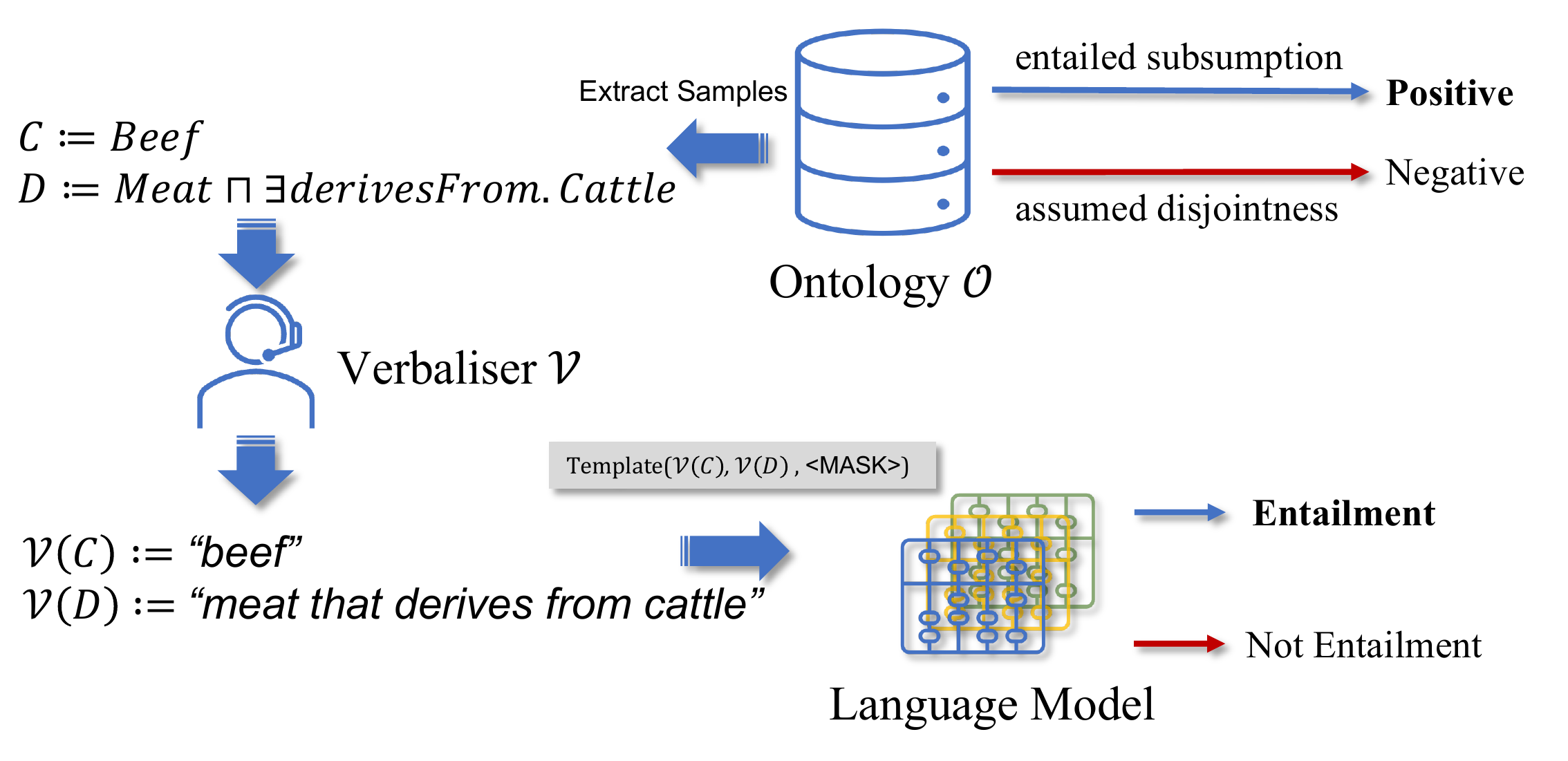}
\caption{\textsc{OntoLAMA} framework.}
\label{fig:ontolama_framework}
\end{center}
\vspace{-2mm}
\end{figure}

In this work, we take a further step along the ``LMs-as-KBs'' research line towards more formalised semantics by targeting DL KBs and in particular the OWL ontologies. Current works on LMs concerning ontologies are mostly driven by a target application. \citet{liu2020concept}, \citet{He_Chen_Antonyrajah_Horrocks_2022}, and \citet{Chen2022ContextualSE} apply language model fine-tuning to address ontology curation tasks such as concept insertion and matching, while \citet{Ye2022OntologyenhancedPF} transform ontologies into graphs for data augmentation in few-shot learning. In contrast to these application-driven approaches, we investigate a more fundamental question: \emph{To what extent can LMs infer conceptual knowledge modelled by an ontology?} Particularly, we focus on the subsumption relationships between ontology concepts. As shown in Figure \ref{fig:ontolama_framework}, we first extract concept pairs $(C, D)$ that are deemed as positive ($C$ and $D$ are in a subsumption relationship) and negative ($C$ and $D$ are assumed to be disjoint) samples from an ontology. Note that the sampling procedure is fully automatic with the syntax and semantics of OWL ontology carefully considered. To translate the concepts and especially the ones with complex logical expressions into natural language texts, we develop a recursive concept verbaliser. We formulate the Subsumption Inference (SI) task similarly to the Natural Language Inference (NLI) task and treat the concept pairs as premise-hypothesis pairs \cite{text-entail-book}, which will then be wrapped into a template for generating inputs of LMs.

We have created SI datasets from ontologies of various domains and scales, and conducted extensive experiments. Our results demonstrate that LMs perform better on a typical NLI task than the constructed SI tasks under the zero-shot setting, indicating that LMs encode relatively less background knowledge of ontology subsumptions. However, by providing a small number of samples ($K$-shot settings), the performance on SI is significantly improved. This observation is consistent with the three LMs that are studied in this work.



\section{Background}

\subsection{OWL Ontology} \label{sec:onto}
An OWL ontology is a description logic (DL) knowledge base that consists of the TBox (terminological), ABox (assertional), and RBox (relational) axioms \cite{Krtzsch2012JaN2}. In this work, we focus on the TBox axioms which specify the subsumption relationships between concepts of a domain. A subsumption axiom has the form of $C \sqsubseteq D$ where $C$ and $D$ are concept expressions involving atomic concept, negation ($\neg$), conjunction ($\sqcap$), disjunction ($\sqcup$), existential restriction ($\exists r.C$), universal restriction ($\forall r. C$), and so on (see complete definition in Appendix \ref{appendix:owl-onto}). An \textbf{atomic concept} is a named concept, a top concept $\top$ (a concept with every individual as an instance), or a bottom concept $\bot$ (an empty concept); while a \textbf{complex concept} consists of at least one of the available logical operators. An equivalence axiom $C \equiv D$ is equivalent to $C \sqsubseteq D$ and $D \sqsubseteq C$. 


Regarding the semantics, in DL we define
an \textit{interpretation} $\mathcal{I} = ({\Delta}^{\mathcal{I}}, \cdot^{\mathcal{I}})$ that consists of an non-empty set ${\Delta}^{\mathcal{I}}$ and a function $\cdot^{\mathcal{I}}$ that maps each concept $C$ to $C^{\mathcal{I}} \subseteq {\Delta}^{\mathcal{I}}$ and each \textit{property} $r$ to $r^{\mathcal{I}} \subseteq {\Delta}^{\mathcal{I}} \times {\Delta}^{\mathcal{I}}$. We say $\mathcal{I}$ is a model of $C \sqsubseteq{D}$ if $C^{\mathcal{I}} \subseteq D^{\mathcal{I}}$ holds, and $\mathcal{I}$ is a model of an ontology $\mathcal{O}$ if $\mathcal{I}$ is a model of all axioms in $\mathcal{O}$. If $C^{\mathcal{I}} \subseteq D^{\mathcal{I}}$ holds for every model $\mathcal{I}$ of $\mathcal{O}$, then we can say $\mathcal{O} \models C \sqsubseteq D$. This defines logical entailment w.r.t.\ an ontology and it is more strictly defined than textual entailment based on human beliefs. 

An individual $a$ is an instance of a concept $C$ in $\mathcal{O}$ if $\mathcal{O} \models C(a)$ ($\inter{a} \in \inter{C}$ for every model $\mathcal{I}$ of $\mathcal{O}$). $C$ and $D$ are \textit{disjoint} in $\mathcal{O}$ if $O \models C \sqcap D \sqsubseteq \bot$ (or equivalently $O \models C \sqsubseteq \neg D$) which means there can be no common instance $a$ of $C$ and $D$.

The Open World Assumption (OWA) underpins OWL ontologies, according to which we cannot say what is not entailed by the ontology is necessarily false. For example, if we have an ontology that contains just one axiom $Paella \sqsubseteq \exists hasIngredient.Chicken$, in OWA we cannot determine if paella can have chorizo as an ingredient or not. To allow reuse and extension, ontologies are often (intentionally) underspecified \cite{Cimiano2003OntologybasedSC}; this characteristic motivates how we define the negative samples in Section \ref{sec:SI-general}.

\subsection{Related Work}

Recently, the rise of the prompt learning paradigm has shed light on better usage of pre-trained LMs without, or with minor, supervision \cite{Liu2022PretrainPA}. However, LMs are typically pre-trained in a stochastic manner, making it challenging to study what knowledge LMs have implicitly encoded \cite{petroni-etal-2019-language} and how to access LMs in an optimal or controllable way \cite{gao-etal-2021-making,Li2022DiffusionLMIC}. 

Our work is informed by the ``LMs-as-KBs'' literature \cite{AlKhamissi2022ARO}, where different probes have been designed to test LMs' knowledge of relational data. In \citet{petroni-etal-2019-language}, the probing task of world knowledge has been formulated as a cloze-style answering task where LMs are required to fill in the \texttt{<MASK>} token given input texts wrapped into a manually designed template. \citet{sung-etal-2021-language} did a similar work but shift the focus to (biomedical) domain knowledge of domain-specific LMs. \citet{Liu2021EnhancingML} pre-trained LMs with multi-lingual knowledge graphs (KGs) and test on the cross-lingual tasks. 
\citet{dhingra-etal-2022-time} proposed datasets with temporal signals and probed LMs on them with templates generated by the text-to-text transformer T5 \cite{T5}.   

However, existing ``LMs-as-KBs'' works mostly focus on relational facts,
but omit logical semantics and conceptual knowledge. In contrast, our work focuses on OWL ontologies which represent conceptual knowledge with an underlying logical formalism.
Although there are some recent works concerning both LMs and ontologies, they do not compare them at the semantic level but rather emphasise on downstream applications. For example, \citet{He_Chen_Antonyrajah_Horrocks_2022} adopted LMs as synonym classifiers to predict mappings between ontologies; whereas \citet{Ye2022OntologyenhancedPF} used ontologies to provide extra contexts to help LMs to make predictions.

\section{Subsumption Inference} \label{sec:SI}

\subsection{Task Definition} \label{sec:SI-general}

Recall the definitions in Section \ref{sec:onto}, a subsumption axiom $C \sqsubseteq D$ can be interpreted as: \state{every instance of $C$ is an instance of $D$}. We can accordingly form a premise-hypothesis pair where the \textit{premise} is \state{$x$ is a $C$} and the \textit{hypothesis} is \state{$x$ is a $D$} for some individual $x$. Note that there are different ways to express the premise and hypothesis, and we adopt a simple but effective one
(see Section \ref{sec:prompt-infer}). Next, an \textit{ontology verbaliser} is required for transforming the concept expressions $C$ and $D$ into natural language texts.
Analogous to Natural Language Inference (NLI) or Recognising Textual Entailment (RTE) \cite{poliak-2020-survey,text-entail-book}, the task of Subsumption Inference (SI) is thus defined as \textit{classifying if the premise entails or does not entail the hypothesis}. Note that SI is similar to a two-way RTE task\footnote{RTE guidelines: \url{https://tac.nist.gov/2008/rte/rte.08.guidelines.html}.}  where we do not consider the \textit{neutral}\footnote{\textit{Neutral} essentially means two terms are unrelated. Ontologies are invariably underspecified, so even if two concepts have not been entailed as a subsumption or non-subsumption, they may still be implicitly related in the real world.} class. 

Given an ontology $\mathcal{O}$, we extract positive and negative subsumptions to probe LMs. The positive samples are concept pairs $(C, D)$ with $\mathcal{O} \models C \sqsubseteq D$. Due to OWA, we cannot determine if $(C, D)$ with $\mathcal{O} \not\models C \sqsubseteq D$ really forms a negative subsumption (see Appendix \ref{appendix:explain-subs} for more explanation); to generate plausible negative samples, we propose the assumed disjointness\footnote{\citet{DBLP:conf/esws/Schlobach05} and \citet{DBLP:journals/kais/SolimandoJG17} defined a similar assumption but in different contexts.} defined as follows:
\begin{definition*}
    If two concepts $C$ and $D$ are satisfiable in $\mathcal{O} \cup \{ C \sqcap D \sqsubseteq \bot \}$ and there is no named atomic concept $A$ in $\mathcal{O}$ such that $\mathcal{O} \models A \sqsubseteq C$ and $\mathcal{O} \models A \sqsubseteq D$, then $C$ and $D$ are assumed to be disjoint.
\end{definition*}
\noindent The first condition ensures that $C$ and $D$ are still \textbf{satisfiable} after adding the disjointness axiom for them into $\mathcal{O}$ whereas the second condition ensures that $C$ and $D$ have \textbf{no common descendants} because otherwise the disjointness axiom will make any common descendant unsatisfiable. If two concepts $C$ and $D$ satisfy these two conditions, we treat $(C, D)$ as a valid negative subsumption.

However, in practice validating the satisfiability for each concept pair $(C, D)$ would be inefficient especially when the ontology is large and complex. Thus, we propose a pragmatical alternative to the satisfiability check in Appendix \ref{appendix:assumed-disjointess-alternative}.

To conduct reasoning to extract entailed positive subsumptions and validate sampled negative subsumptions, we need to adopt a proven sound and complete OWL reasoner, e.g., HermiT \cite{hermit}.

In the following sub-sections, we propose two specific SI tasks and their respective subsumption sampling methods.

\subsection{Atomic Subsumption Inference} \label{sec:atom_si}
The first task aims at subsumption axioms that involve just \textit{named atomic concepts}. Such axioms are usually the most prevalent in an ontology and can be easily verbalised by using the concept names. In this work, we use labels (in English) defined by the built-in annotation property \texttt{rdfs:label} as concept names.

The positive samples are extracted from all entailed subsumption axioms of the target ontology. 
We consider two types of negative samples: \textit{(i)} \textbf{soft negative} composed of two random concepts, and \textit{(ii)} \textbf{hard negative} composed of two random \textit{sibling} concepts. Two sibling concepts lead to a ``hard'' negative sample because they share a common parent (thus having closer semantics) but are often disjoint. The sampled pairs need to meet the assumed disjointness defined in Section \ref{sec:SI-general} to be accepted as valid negatives. 
We first sample equal numbers of soft and hard negatives and then randomly truncate the resulting set into the size of the positive sample set to keep class balance.

\subsection{Complex Subsumption Inference} \label{sec:comp_si}

\begin{table}[!t]
\centering
\renewcommand{\arraystretch}{1.3} 
{\fontsize{9}{10.6}\selectfont
\begin{tabularx}{0.49\textwidth}{l X} \toprule
     \textbf{Pattern} & \textbf{Verbalisation ($\mathcal{V}$)} \\\midrule
     $A$ (atomic) & the name (\texttt{rdfs:label}) of $A$  \\
     $r$ (property) & the name (\texttt{rdfs:label}) of $r$, subject to rules in Appendix \ref{appendix:object-prop} \\
     $\neg C$ & \textit{``not $\mathcal{V}(C)$''} \\
     $\exists r.C$ & \textit{``something that $\mathcal{V}(r)$ some $\mathcal{V}(C)$''} \\
     $\forall r.C$ & \textit{``something that $\mathcal{V}(r)$ only $\mathcal{V}(C)$''} \\
     $C_1 \sqcap ... \sqcap C_n$ & 
     if $C_i = \exists/\forall r.D_i$ and $C_j = \exists/\forall r.D_j$, they will be re-written into $\exists/\forall r.(D_i \sqcap D_j)$ before verbalisation; suppose after re-writing the new expression is $C_1 \sqcap ... \sqcap C_{n'}$ 
     \newline
     \textbf{(a)} if \textbf{all} $C_i$s (for $i = 1, ..., n'$) are restrictions, in the form of $\exists/\forall r_i.D_i$: \newline
     \textit{``something that $\mathcal{V}(r_1)$ some/only $V(D_1)$ and ... and $\mathcal{V}(r_{n'})$ some/only $V(D_{n'})$''} \newline
     \textbf{(b)} if \textbf{some} $C_i$s (for $i = m+1, ..., n'$) are restrictions, in the form of $\exists/\forall r_i.D_i$: \newline
     \textit{``$\mathcal{V}(C_{1})$ and ... and $\mathcal{V}(C_{m})$ that $\mathcal{V}(r_{m+1})$ some/only $V(D_{m+1})$ and ... and $\mathcal{V}(r_{n'})$ some/only $V(D_{n'})$''} \newline
     \textbf{(c)} if \textbf{no} $C_i$ is a restriction: \newline
     \textit{``$\mathcal{V}(C_{1})$ and ... and $\mathcal{V}(C_{n'})$''}
     \\
     $C_1 \sqcup ... \sqcup C_n$ & similar to verbalising $C_1 \sqcap ... \sqcap C_n$ except that \textit{``and''} is replaced by \textit{``or''} and case \textbf{(b)} uses the same verbalisation as case \textbf{(c)}
     \\\bottomrule
\end{tabularx}}
\caption{Recursive rules for verbalising a complex concept expression $C$ in OWL ontologies. Note that $C_i$ in the conjunction/disjunction pattern is also an arbitrary complex concept.}
\label{tab:verbalisation-rules}
\vspace{-3mm}
\end{table}

\begin{figure}[t!]
\begin{center}
    \includegraphics[width=0.48\textwidth]{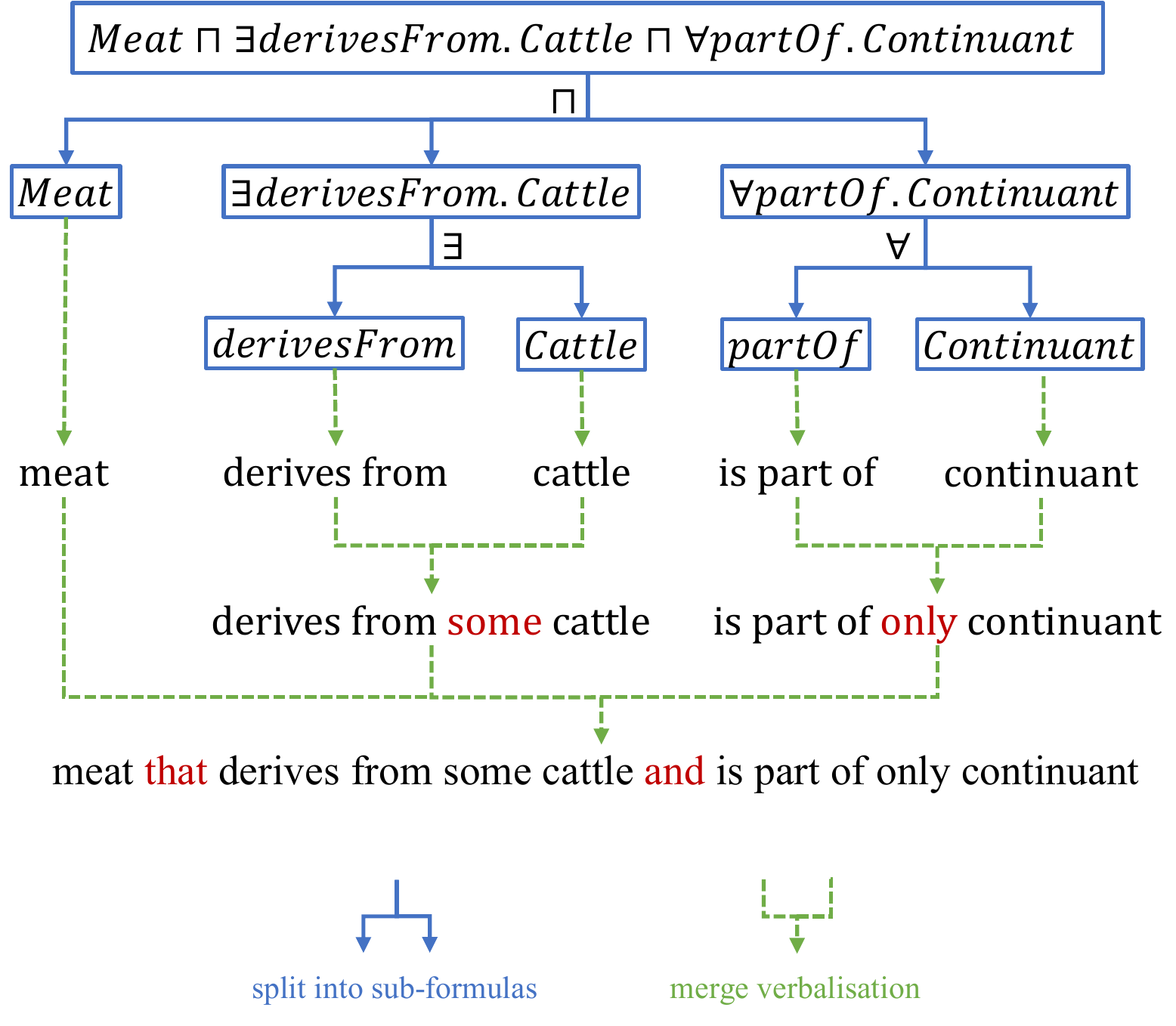}
\caption{Illustration of how the recursive concept verbaliser is applied to an example complex concept expression. The algorithm first {\color{pptblue} splits} the complex concept into a sub-formula tree, verbalising the leaf nodes, and then {\color{pptgreen} merging} the verbalised sub-formulas recursively. The key word associated with the logical operator at each merging step is marked in {\color{pptred} red}. See Appendix \ref{appendix:data-examples} for more examples.}
\label{fig:verbalisation-example}
\end{center}
\vspace{-4mm}
\end{figure}

In the second SI task, we consider subsumption axioms that involve \textit{complex concepts}. Particularly, we choose equivalence axioms of the form $A \equiv C$\footnote{Equivalence axioms of this form are referred to as the \textit{definition} of the named concept, and are common in OWL.} (where $A$ and $C$ are atomic and complex concepts, respectively) as anchors, and equivalently transform them into subsumption axioms of the forms $A \sqsubseteq C$ and $C \sqsubseteq A$, through which complex concepts can appear on both the premise and hypothesis sides.


\subsubsection*{Recursive Concept Verbaliser} 

To transform a complex $C$ into a natural language text, we develop the \textit{recursive concept verbaliser} consisting of a \textit{syntax tree parser} and a set of \textit{recursive rules} (see Table \ref{tab:verbalisation-rules}). A concrete example is shown in Figure \ref{fig:verbalisation-example}, where the complex concept  $Meat \sqcap \exists derivesFrom.Cattle \sqcap \forall partOf.Continuant$ is first split into a sub-formula tree by the syntax parser, then verbalised according to the recursive rules in Table \ref{tab:verbalisation-rules}. The leaf nodes are either atomic concepts or properties and they are verbalised by their names. At each recursive step, verbalised child nodes are merged according to the logical pattern in their parent node. Note that we enforce some extent of redundancy removal for the conjunction ($\sqcap$) and the disjunction ($\sqcup$) patterns. Taking the example in Figure \ref{fig:verbalisation-example}, the verbalised atomic concept \textit{``meat''} is placed before \textit{``that''} as an \textit{antecedent}, and the verbalised conjunction of two restrictions is placed after \textit{``that''} as a \textit{relative clause}. \textit{``meat''} can be replaced by \textit{``something''} if the concept $Meat$ is not involved. 
Moreover, if two restrictions with the same quantifier and property are connected by $\sqcap$ or $\sqcup$, they will be merged into one restriction. For example, $\exists derivesFrom.Cattle \sqcap \exists derivesFrom.Sheep$ will be transformed into $\exists derivesFrom.(Cattle \sqcap Sheep)$. 


\begin{table*}[!t]
\centering
\renewcommand{\arraystretch}{1.1} 
{\fontsize{9}{10.8}\selectfont
\begin{tabularx}{0.83\textwidth}{l c c c X} \toprule
     \textbf{Source} & \textbf{\#Concepts} & \textbf{\#EquivAxioms} & \phantom{a} & \textbf{\#Dataset (Train/Dev/Test)}  \\\midrule
    \schema & $894$ & - & & Atomic SI: $\ 808/404/2,830$ \\\midrule
    \doid & $11,157$ & - & & Atomic SI: $\ 90,500/11,312/11,314 $ \\\midrule
    \multirow{2}{5em}{\foodon} & \multirow{2}{5em}{\centering $30,995$} & \multirow{2}{5em}{\centering $2,383$} & & \multirow{2}{16em}{Atomic SI: $\ 768,486/96,060/96,062$ \newline Complex SI: $\ 3,754/1,850/13,080$} \\\\\midrule
    \multirow{2}{5em}{\go} & \multirow{2}{5em}{\centering $43,303$} &  \multirow{2}{5em}{\centering $11,456$} & & \multirow{2}{16em}{Atomic SI: $\ 772,870/96,608/96,610$ \newline Complex SI: $\ 72,318/9,040/9,040$}\\\\\midrule
    \texttt{MNLI} & - & - & & biMNLI: $\ 235,622/26,180/12,906 $ \\
     \bottomrule
\end{tabularx}}
\caption{Statistics for ontologies, SI datasets, and the biMNLI dataset.}
\label{tab:data}
\end{table*}


We extract equivalence axioms in the form of $A \equiv C$ from the target ontology. Taking each such axiom as an anchor, we can obtain positive complex subsumption axioms of the form $A_{sub} \sqsubseteq C$ or $C \sqsubseteq A_{super}$ where $A_{sub}$ and $A_{super}$ are a sub-class and a super-class of $A$, respectively. To derive challenging negative samples, we first randomly replace a named concept or a property in $A \equiv C$ to generate either \textit{(i)} $A' \equiv C$ (if $A$ is replaced by $A'$) or \textit{(ii)} $A \equiv C'$ (if $C$ is corrupted). Without loss of generality, we assume the random replacement leads to case~\textit{(ii)}. We then check if $A$ and $C'$ satisfy the assumed disjointness as described in Section~\ref{sec:SI-general}. In the affirmative case, 
we can have either $A \sqsubseteq C'$ or $C' \sqsubseteq A$ as the final negative subsumption; otherwise, we skip this sample. 
For example, given $SunflowerSeed \equiv Seed \ \sqcap \ \exists DerivesFrom.HelianthusAnnuus$, a possible negative subsumption is $SunflowerSeed \sqsubseteq Fruit \ \sqcap \ \exists DerivesFrom.HelianthusAnnuus$ if $Seed$ in $C$ is replaced by $Fruit$ to create~$C'$.


\section{Datasets} \label{sec:datasets}

In this work, we consider ontologies of different domains and scales including:
\begin{itemize}
    \item \schema\footnote{\url{https://schema.org/}} (released on \texttt{2022-03-17}): a general-purpose ontology that maintains a basic schema for structured data on the Web;
    
    \item \doid\footnote{\url{https://disease-ontology.org/}} (released on \texttt{2022-09-29}): an ontology for human diseases \cite{Schriml2012DiseaseOA};
    
    \item
    \foodon\footnote{\url{https://foodon.org/}} (released on \texttt{2022-08-12}): an ontology specialised in food-related knowledge including food products, food sources, food nutrition, and so on \cite{Dooley2018FoodOnAH}.

    \item \go\footnote{\url{http://geneontology.org/}} (released on \texttt{2022-11-03}): a very fine-grained and widely used biomedical ontology specialised in genes and gene functions \cite{Ashburner2000GeneOT}.
\end{itemize}
We used the most updated versions at the time of experiment. The details for pre-processing the ontologies are illustrated in Appendix~\ref{appendix:onto_preprocess}.

We construct an Atomic SI dataset for each ontology, but Complex SI datasets are created for \foodon and \go only, due to their abundance of equivalence axioms. 
To avoid too many repetitive concept expressions brought by a particular equivalence axiom, we sample at most $4$ positive and $4$ negative samples for each equivalence axiom in the Complex SI setting.
To attain class balance, we purposely keep the number of negative samples the same as the positive samples in each data split. For most of the resulting datasets, we divide each into $8:1:1$ for training, development, and testing; 
for the \schema's Atomic SI and the \foodon's Complex SI datasets, which are relatively smaller,
we apply a $2:1:7$ division instead. Note that we mainly focus on $K$-shot settings in the probing study, thus the required training and development sample sets are small.

To compare with how LMs perform on traditional NLI, we additionally create biMNLI, a subset of the Multi-Genre Natural Language Inference (\texttt{MNLI}) corpus \cite{williams-etal-2018-broad} where \textit{(i)} the neutral class and its samples are removed, \textit{(ii)} the Matched and Mismatched testing sets are merged into one testing set, \textit{(iii)} $10\%$ of the training data is used as the development set, and \textit{(iv)} the entailment-contradiction ratio is set to $1:1$ (by discarding extra samples from the dominant class) for a balanced prior. The numbers of named concepts and equivalence axioms in ontologies, and the numbers of samples in (each split of) SI datasets and the biMNLI dataset are reported in Table \ref{tab:data}.


\section{Experiments}

\subsection{Prompt-based Inference} \label{sec:prompt-infer}

To conduct the inference task under the prompt-based settings, we wrap the verbalised subsumption axioms and the \texttt{<MASK>} token into a template to serve as inputs of LMs. We opt to use different combinations of manually designed templates\footnote{We make slight modifications by adding the prefix ``It/it is \texttt{<A>}'' to make premise and hypothesis sentences complete.} ($T_1$ and $T_2$) and label words ($L_1$ to $L_3$) that have achieved promising results on the NLI tasks \cite{schick-schutze-2021-exploiting,gao-etal-2021-making} as follows:
\vspace{-.4cm}

{\fontsize{10}{12}\selectfont
\begin{align*}
    T_1 := & \ \underbrace{\text{It is \texttt{<A>} } \mathcal{V}(C)}_{\text{premise}}\text{? \texttt{<MASK>}, } \underbrace{\text{ it is \texttt{<A>} } \mathcal{V}(D)}_{\text{hypothesis}}\text{.} \\
    T_2 := & \ \text{``}\underbrace{\text{It is \texttt{<A>} } \mathcal{V}(C)}_{\text{premise}}\text{''? \texttt{<MASK>}, ``} \underbrace{\text{it is \texttt{<A>} } \mathcal{V}(D)}_{\text{hypothesis}}\text{''.} 
\end{align*}
\begin{align*}
    L_1 := &\text{ \{``positive'': [``Yes''], ``negative'': [``No'']\}} \\
    L_2 := &\text{ \{``positive'': [``Right''], ``negative'': [``Wrong'']\}} \\
    L_3 := &\text{ \{``positive'': [``Yes'', ``Right''],} \\
    &\text{ \phantom{a}``negative'': [``No'', ``Wrong'']\}}
\end{align*}}

\noindent where \texttt{<A>} is \state{a}, \state{an}, or just \textit{blank} depending on the next word\footnote{``an'' is used when the next word starts with a vowel; leaving it blank when the next word is \textit{``something''}.}, $\mathcal{V}(\cdot)$ is the concept verbalisation function defined in Section \ref{sec:SI}, and \texttt{<MASK>} is the token that LMs need to predict. The probability of predicting class $y$ (``positive'' or ``negative'') for an input sample $x = (C, D)$ is defined as:
\begin{align*}
    P(y \ | \ x) &= P(\texttt{<MASK>} \in L_j[y]) \ | \ T_i(C, D)) \\
    &= \frac{\sum_{v \in L_j[y]} \exp(\mathbf{w}_v \cdot \mathbf{h}_{\texttt{<MASK>}})}{\sum_{w \in L_j[\cdot]} \exp(\mathbf{w}_w \cdot \mathbf{h}_{\texttt{<MASK>}})} 
\end{align*} 
where $L_j[\cdot]$ and $L_j[y]$ denote all the label words defined in $L_j$ and the label words of class $y$ defined in $L_j$, respectively; $T_i(C, D)$ denotes the transformed texts of concepts $C$ and $D$ through the template $T_i$; $\mathbf{w}_{v}$ and $\mathbf{w}_{w}$ are vectors for the label words $v$ and $w$, respectively; and $\mathbf{h}_{\texttt{<MASK>}}$ denotes the hidden vector of the masked token. The prediction can be trained by minimising the cross-entropy loss.

For the biMNLI dataset, the premise and hypothesis are replaced by what were originally given in the dataset -- except that we have removed trailing punctuations.

\begin{table*}
    \centering
    \renewcommand{\arraystretch}{1.2} 
    {\fontsize{9}{10.8}\selectfont
    \begin{tabularx}{0.9\textwidth}{X Y Y Y Y Y Y Y}
    \toprule
     & & \multicolumn{4}{c}{\textbf{Atomic SI}} & \multicolumn{2}{c}{\textbf{Complex SI}}\\
    \cmidrule(lr){3-6} \cmidrule(lr){7-8}
    \textbf{Setting} & \textbf{biMNLI} &  \textbf{Schema.org} & \textbf{DOID} & \textbf{FoodOn} & \textbf{GO} & \textbf{FoodOn} & \textbf{GO}\\
    \midrule
    \textsf{majority} & 50.0 &  50.0 &  50.0 &  50.0 & 50.0  &  50.0 & 50.0 \\ 
    \multicolumn{2}{l}{\textsf{word2vec}} \\
    \hspace{3mm} K=4 & 51.5 (0.2) & 54.9 (2.9) & 64.6 (2.6) & 63.5 (1.0) & 60.1 (4.1)  & 56.8 (4.2) & 56.9 (5.4) \\
    \hspace{3mm} K=128 & 52.1 (0.4) & 73.0 (0.4) & 70.8 (1.7) & 71.4 (1.0) & 66.3 (0.9)  & 63.8 (0.6) & 66.4 (1.3) \\ \midrule
    \multicolumn{2}{l}{\textsf{roberta-base}} \\
    \hspace{3mm} K=0   &  62.5 (6.5) &  56.4 (3.6) &  53.3 (4.0) &  54.6 (4.4) &  49.0 (2.4) &    55.9 (3.6) &  48.7 (3.1) \\
    \hspace{3mm} K=4   &  67.6 (5.2) &  62.9 (5.2) &  61.8 (6.7) &  62.1 (4.2) &  65.2 (5.0) &    62.4 (3.2) &  52.2 (7.1) \\
    \hspace{3mm} K=32  &  78.8 (1.1) &  84.3 (2.0) &  89.0 (1.4) &  85.0 (1.1) &  84.6 (2.5) &    77.0 (1.5) &  76.4 (2.5) \\
    \hspace{3mm} K=128 &  85.1 (1.0) &  91.1 (0.7) &  92.4 (0.7) &  90.0 (0.7) &  89.0 (0.8) &    85.5 (1.3) &  86.9 (1.5) \\
    \midrule
    \multicolumn{2}{l}{\textsf{roberta-large}} \\
    \hspace{3mm} K=0   &  68.7 (6.2) &  61.7 (7.2) &  59.8 (5.4) &  60.1 (8.8) &  54.6 (1.9) &    56.1 (1.9) &  50.4 (0.6) \\
    \hspace{3mm} K=4   &  78.1 (6.6) &  69.4 (5.4) &  74.0 (5.5) &  71.6 (4.4) &  67.6 (3.4) &    64.1 (5.1) &  56.9 (5.7) \\
    \hspace{3mm} K=32  &  89.9 (1.2) &  87.3 (1.9) &  92.3 (0.7) &  88.9 (1.6) &  87.7 (1.6) &    80.8 (3.8) &  81.6 (2.2) \\
    \hspace{3mm} K=128 &  93.0 (0.8) &  92.9 (0.8) &  93.4 (0.5) &  92.2 (0.5) &  91.0 (0.7) &    88.4 (1.1) &  90.2 (1.0) \\
    \cmidrule(lr){1-8}
    \hspace{3mm} full  &  97.5 &  95.4  &  97.8  &  98.7  &  98.1  &    95.8  &  98.8  \\
    \bottomrule
    \end{tabularx}}
    \caption{Results for the biMNLI, Atomic SI, and Complex SI tasks with each cell stating \textit{``mean accuracy (standard deviation)''} except for \textsf{majority vote} and the fully supervised settings where standard deviation is not available.}
    \label{tab:main_exps}
    \vspace{-2mm}
\end{table*}

\begin{figure*}[t!]
\begin{center}
    \includegraphics[width=0.8\textwidth]{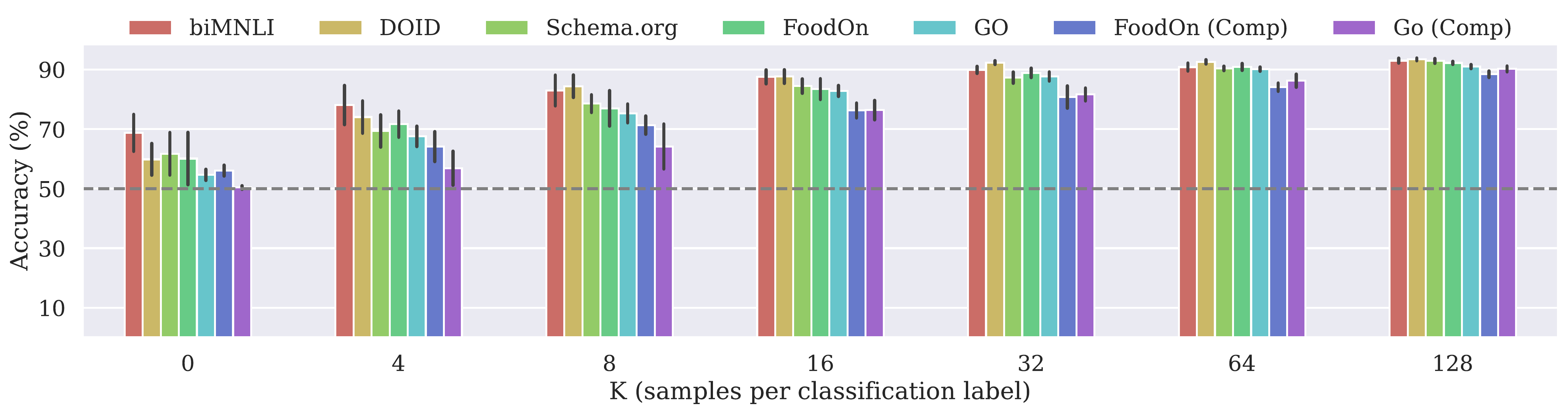}
\caption{Visualisation of K-shot results (for \textsf{roberta-large}) on the biMNLI, Atomic SI, and Complex SI tasks, where the dotted horizontal line indicates \textsf{majority vote}. The order of the bars is the same as in the legend.}
\label{fig:roberta-large}
\end{center}
\vspace{-4mm}
\end{figure*}

In the main experiments concerning language models, we consider all the combinations of $T_i$ and $L_j$ and additionally consider $3$ random seeds (thus $18$ experiments each) for $K$-shot settings where $K > 0$. The value of $K$ refers to the number of samples per classification label (positive or negative) we randomly extract from training and development sets, respectively. For $K=0$ (zero-shot), different random seeds do not affect the results. For the fully supervised setting, we consider only one random seed and one combination ($T_1$ and $L_1$) because our pilot experiments demonstrate that fine-tuning on large samples results in low variance brought by different random seeds and different combinations of templates and label words.

Our code implementations mainly rely on \texttt{The~OWL~API}\footnote{\url{https://owlapi.sourceforge.net/}} for ontology processing and reasoning, and \texttt{OpenPrompt}\footnote{\url{https://thunlp.github.io/OpenPrompt/}} for prompt learning \cite{ding-etal-2022-openprompt}. Training of each $K$-shot (where $K>0$) experiment takes 10 epochs, while for the fully supervised setting involving very large training samples, we only train for 1 epoch.\footnote{Since \schema's Atomic SI and \foodon's Complex SI datasets have a small training set, their fully supervised settings still take 10 epochs.} The best-performing model on the development set (at each epoch) is selected for testing set inference. We use the AdamW optimiser \cite{Loshchilov2019DecoupledWD} with the initial learning rate, weight decay, and the number of warm-up steps set to $10^{-5}$, $10^{-2}$, and $50$, respectively.
All our experiments are conducted on two \texttt{Quadro RTX 8000} GPUs. 

\subsection{Results and Analysis}

\subsubsection*{LMs and Settings} We choose LMs from the RoBERTa family \cite{liu2019roberta} as they are frequently introduced in cloze-style probing tasks \cite{liu-etal-2021-probing-across,sung-etal-2021-language,kavumba-etal-2022-prompt}. In Table \ref{tab:main_exps}, we present key experiment results for \textsf{roberta-large} and \textsf{roberta-base}; we have a further ablation study for a biomedical variant of \textsf{roberta-large} in the latter paragraph.

For both LMs in Table \ref{tab:main_exps}, we report results of $K$-shot settings with $K \in \{0,4,32,128\}$. We additionally present the results of the fully supervised setting for \textsf{roberta-large} as the oracle. For each setting, we report the averaged accuracy and standard deviation (where applicable). To clearly observe how the performance varies as $K$ increases, we present Figure \ref{fig:roberta-large} which visualises the $K$-shot results for \textsf{roberta-large} with additional values of $K$ ($\{8,16,64\}$). The complete result table for both language models and the figure that visualises the performance of \textsf{roberta-base} are available in Appendix \ref{appendix:complement}.

\subsubsection*{Baselines}
As aforementioned, we purposely keep class balance in each data split, thus the accuracy scores for \textsf{majority vote} are all $50.0\%$. Besides, we consider \textsf{word2vec} \cite{mikolov2013efficient} pre-trained on GoogleNews\footnote{\url{https://code.google.com/archive/p/word2vec/}} with a logistic regression classifer as a baseline model, which demonstrates how a classic non-contextual word embedding model performs on the SI tasks. For this baseline, we only report results for $K \in \{4, 128\}$ as the increase of $K$ does not bring significant change and results of $K=128$ are roughly comparable to results of $K=4$ for \textsf{roberta-large}. This suggests that the SI sample patterns are not easily captured with \textsf{word2vec}. 

\subsubsection*{SI vs biMNLI
}
From the results, we first observe that both \textsf{roberta-large} and \textsf{roberta-base} achieve better zero-shot results on biMNLI than on all the SI datasets by at least $7.0\%$ and $6.1\%$ respectively, showing that under our prompt settings, both LMs encode better background knowledge on biMNLI than SI. However, as $K$ grows, the performances on both biMNLI and SI improve consistently and significantly (while the standard deviation generally reduces), and we can see at $K=32$, the mean accuracy scores on the Atomic SI tasks have surpassed biMNLI for \textsf{roberta-base}. At $K=64$ (see Figure \ref{fig:roberta-large}), the mean accuracy scores on biMNLI and all the Atomic SI tasks converge to around $90.0\%$; the scores on the two Complex SI tasks are also above $80.0\%$ for both LMs.
Moreover, \textsf{roberta-large} consistently attains a better score than \textsf{roberta-base} for every setting.


\subsubsection*{Comparison Among SI Tasks}
We observe that Complex SI is generally harder than Atomic SI. For example, at $K=0$, \textsf{roberta-large} attains $50.4\%$ almost as \textsf{majority vote} on the Complex SI dataset of \go; at $K=128$, \textsf{roberta-large} attains $88.4\%$ on the Complex SI dataset of \foodon while it attains more than $90\%$ for the others. We can also observe from Figure \ref{fig:roberta-large} that the scores on Complex SI tasks are generally lower than those on the Atomic SI tasks. Among the Atomic SI tasks, we find that \go is the most challenging which is as expected because \go is a fine-grained expert-level ontology. However, it surprises us that at $K=32$ the score ($92.3\%$) on \doid is better than all other tasks, considering that \doid is a domain-specific ontology.

\subsubsection*{Domain-specific SI}
We conduct further experiments for domain-specific LMs on domain-specific SI tasks. Specifically, we consider the variant \textsf{roberta-large-pm-m3-voc} which has been pre-trained on biomedical corpora PubMed abstracts, PMC full-text, and MIMIC-III clinical notes with an updated sub-word vocabulary learnt from  PubMed \cite{lewis-etal-2020-pretrained}. In Table \ref{tab:bio-roberta}, we present the $K$-shot results of \textsf{roberta-large-pm-m3-voc} on three SI tasks related to biomedical ontologies \doid and \go. The zero-shot scores are almost equivalent to \textsf{majority vote} but the performance improves more prominently than \textsf{roberta-large} on the Atomic SI tasks of \doid and \go as $K$ increases. Surprisingly, the Complex SI setting of \go seems to be quite challenging to this biomedical variant of RoBERTa. For example, at $K=4$, the score is not improved compared to $K=0$.


\renewcommand{\arraystretch}{1.05} 
\begin{table}[t!]
    \centering
    {\fontsize{9}{10.8}\selectfont
    \begin{tabularx}{0.39\textwidth}{l c c c}
    \toprule
    \textbf{K} & \textbf{DOID} &          \textbf{GO} &   \textbf{GO (Comp)} \\
    \midrule
    0   &  49.7 (0.4) &  50.1 (0.2) & 50.0 (0.0) \\
    4   &  64.8 (7.9) &  66.2 (6.5) &  50.0 (0.7) \\
    32  &  94.7 (1.3) &  93.5 (1.1) &  73.5 (3.6) \\
    128 &  96.3 (0.4) &  95.2 (0.5) &  90.5 (1.8) \\
    \bottomrule
    \end{tabularx}}
    \caption{Results for \textsf{roberta-large-pm-m3-voc} on SI tasks of biomedical ontologies \doid and \go.}
    \label{tab:bio-roberta}
    \vspace{-4mm}
\end{table}

\subsubsection*{Template and Label Words}

The access to LMs is an influential factor of performance especially when there are no or fewer training samples. For example, \textsf{roberta-large} attains a standard deviation of $8.8\%$ for $K=0$ on \foodon's Atomic SI task, suggesting that there is a significant performance fluctuation brought by different combinations of templates and label words. Although the standard deviation on \go's Complex SI is just $0.6\%$, the corresponding accuracy score ($50.4\%$) indicates that none of these combinations work. Furthermore, effective template or label words are not transferable from one LM to another, as we can observe from the bad performance of \textsf{roberta-large-pm-m3-voc} for $K=0$ on the SI tasks of biomedical ontologies. These observations suggest that either we did not find a generalised template and label words combination, or LMs require customised access for different types of knowledge.

\section{Conclusion and Discussion}



As a 
work that introduces ontologies to the ``LMs-as-KBs'' collection, this paper emphasises on how to establish a meaningful adaptation from logical expressions to natural language expressions, following their formal semantics.
To this end, we leverage the Natural Language Inference (NLI) setting to define the Subsumption Inference (SI) task with careful considerations to address the differences between textual entailment and logical entailment. 
We also develop the recursive concept verbaliser for OWL ontologies as an auxiliary tool.
Our results demonstrate that with our SI set-ups, LMs can successfully learn to infer both atomic and complex subsumptions when a small number of annotated samples are provided. This paves the way for investigating more complex reasoning tasks
with LMs or guiding LMs using ontology semantics 
with limited
training.

In fact, the current SI setting is not the only way 
for probing subsumption knowledge of an ontology; for example we can directly verbalise $C \sqsubseteq D$ as \state{$\mathcal{V}(C)$ is a kind of $\mathcal{V}(D)$} 
and formulate the probing task similar to fact-checking or equivalently, an inference task with empty premises. However, our pilot experiments demonstrate that such setting is not as effective as the current SI setting. 
 
The presented work brings opportunities for future work as 
\textit{(i)} the proposed ontology verbalisation method has not covered all possible patterns of complex concepts (e.g., with 	cardinality restrictions and nominals); \textit{(ii)} we have not fully considered textual information such as synonyms, definitions, and comments, that are potentially available in an ontology; \textit{(iii)} we have considered only TBox (terminological) axioms, but ABox (assertional) axioms can be involved in, e.g., the membership prediction task, where the objective is to classify which concept an individual belongs to.
Therefore, developing a robust tool for verbalising logical expressions and extending the ontology inference settings are potential next tasks. Another interesting line for the near future is to train an LM using ontologies with their logical semantics considered. 
The resulting LM is expected to be applicable to different downstream ontology curation tasks such as ontology matching and entity linking, with fewer samples necessary for fine-tuning.



\section*{Limitations}


As we mainly focus on conceptual knowledge captured in so-called TBox (terminological) axioms, the ABox (assertional) axioms are not considered. ABox axioms can capture situations for specific individuals (e.g., health status of a person) which could cause privacy issue and we would not expect LMs to capture such knowledge. Hence, dealing with ABox axioms could require additional engineering for data preprocessing.

\section*{Ethical Considerations}
In this work, we construct new datasets for the proposed Subsumption Inference (SI) task from publicly available ontologies: \schema, \doid, \foodon, and \go, with their download links specified in Section \ref{sec:datasets}. The biMNLI dataset is constructed from the existing open-source MNLI dataset. We have confirmed that there is no privacy or license issue in all these datasets.

\section*{Acknowledgements}

This work was supported by Samsung Research UK (SRUK), and the EPSRC projects OASIS (EP/S032347/1), UK FIRES (EP/S019111/1) and ConCur (EP/V050869/1).

\bibliography{anthology,custom}
\bibliographystyle{acl_natbib}

\appendix


\section{OWL Ontology Concept Expression} \label{appendix:owl-onto}

The Description Logic $\mathcal{SROIQ}$ underlies the semantics of OWL~2 ontologies. 
Given the top concept $\top$, the bottom concept $\bot$, the named concept $A$, an individual $a$, a role (or property) $r$ and a non-negative integer $n$,  $\mathcal{SROIQ}$ concept expressions are constructed as:
\begin{align*}
    C,D ::=& \top | \bot | A | (C \sqcap D) | (C \sqcup D) | \neg C |  \exists r.C | \\
    &\forall r.C | \geq n \ r.C | \leq n \ r.C | \exists r.Self | \{a\}
\end{align*}
Recall the definition of \textit{interpretation} $I = (\inter{\Delta}, \inter{\cdot})$, where $\inter{\Delta}$ is a non-empty set (the domain) and $\inter{\cdot}$ maps each concept $C$ to $\inter{C} \subseteq \inter{\Delta}$, a each property $r$ to $\inter{r} \subseteq \inter{\Delta} \times \inter{\Delta}$ and each individual $a$ to an element $\inter{a} \in \inter{\Delta}$. We present the semantics of the concept constructors in Table \ref{tab:semantics-owl}.


\section{Ontology Preprocessing} \label{appendix:onto_preprocess}

In case that some of the ontologies we use in this work contain meaningless (e.g., obsolete) concepts regarding subsumption sampling and/or contain concept names (or aliases) that are apparently unnatural, we apply a \textbf{general} preprocessing procedure to all the ontologies, and then conduct \textbf{individual} preprocessing for each ontology.

\subsubsection*{General Preprocessing}
\begin{itemize}
\itemsep0em 
    \item Remove obsolete concepts (which are indicated by the built-in annotation property \texttt{owl:deprecated}) and apparently redundant concepts such as \texttt{foodOn:stupidType}. 
    \item Use \texttt{rdfs:label} as the main annotation property to extract concept names except when its literal value is not available. The extracted concept names are lower-cased and any underscores \state{\_} in them are removed.
\end{itemize}

\renewcommand{\arraystretch}{1.2} 
\begin{table}[t!]
    \centering
    {\fontsize{8}{9.6}\selectfont
    \begin{tabularx}{0.48\textwidth}{c Y} \toprule
    \textbf{Constructor} & \textbf{Semantics} \\
    \midrule
    $A$ & $\inter{A}$ \\
    $C \sqcap D$ & $ \inter{C} \cap \inter{D}$  \\
    $C \sqcup D$ & $ \inter{C} \cup \inter{D}$  \\
    $\neg C$ & $\inter{\Delta} \ \backslash \ \inter{C}$ \\
    $\top$ & $\inter{\Delta}$ \\
    $\bot$ & $\emptyset$ \\
    $\exists r. C$ & $\{ x \ | \text{ some } \inter{r}\text{-successor of } x \text{ is in }\inter{C} \}$ \\
    $\forall r. C$ & $\{ x \ | \text{ all } \inter{r}\text{-successors of } x \text{ are in }\inter{C} \}$ \\
    $\geq n \ r. C$ & $\{ x \ | \text{ at least } n \ \inter{r}\text{-successors of } x \text{ are in }\inter{C} \}$ \\
    $\leq n \ r. C$ & $\{ x \ | \text{ at most } n \ \inter{r}\text{-successors of } x \text{ are in }\inter{C} \}$ \\
    $\exists r.Self$ & $\{ x \ | \ \langle x,x \rangle \in \inter{r} \}$ \\
    $\{a\}$ & $\{ \inter{a} \}$ \\
    \bottomrule
    \end{tabularx}}
    \caption{Semantics of the OWL Ontology concept constructors.}
    \label{tab:semantics-owl}
    \vspace{-4mm}
\end{table}

\begin{table*}[t!]
    \centering
    \renewcommand{\arraystretch}{1.3} 
    {\fontsize{8}{9.6}\selectfont
    \begin{tabularx}{0.98\textwidth}{X X} \toprule
    \textbf{Complex Concept $C$} & \textbf{Verbalisation} $ \mathcal{V}(C)$ \\\midrule
    $BioRegulation \sqcap \exists negRegulate.ProlineBiosynProc$ & \state{biological regulation that negatively regulates some proline biosynthetic process} \\
    $ApoptoticProc \sqcap \exists partOf.Luteolysis$ & \state{apoptotic process that is part of some lutelysis} \\
    $ConcnOf \sqcap \exists charOf.(fucose \sqcap \exists partOf.MaterialEnt)$ & \state{concentration of something that is characteristic of some fucose that is part of some material entity} \\
    $\exists derivesFrom.(TimothyPlant \sqcup TrifoliumPratense) \sqcap PlantFoodProd \sqcap Silage$ & \state{silage and plant food product that derives from some timothy plant or trifolium pratense} \\
    $Apple \sqcap \neg \exists hasPart.ApplePeel$ & \state{apple (whole or parts) and not something that has part some apple peel}
    \\\bottomrule
    \end{tabularx}}
    \caption{Examples of verbalised complex concepts from \go's and \foodon's Complex SI datasets. Note that in the real datasets, the named concepts and object properties are represented by their IRIs (unique identifiers) instead of the abbreviated names shown in the table.}
    \label{tab:data-examples}
    \vspace{-2mm}
\end{table*}

\subsubsection*{Individual Preprocessing}
\begin{itemize}
\itemsep0em 
    \item \schema: concept names (defined in this ontology are in the Java-identifier style; thus, they are parsed into natural expressions, e.g., \textit{``APIReference''} to \textit{``API Reference''}.
    \item \doid: remove the concept \texttt{doid:Disease} because it is a general concept just below the root concept \texttt{owl:Thing} which will lead to too many simple subsumptions in the form of $C \sqsubseteq Disease$.
    \item \foodon: reconstruct label strings containing non-natural-language texts of three regular expression patterns (note that \verb|(.*)| captures what to be preserved):  
    \begin{enumerate}[\itshape(a)]
        \item \verb|[0-9]+ - (.*) \(.+\)|
        \item \verb|\('(.*)\(gs1', 'gpc\)'\)|
        \item \verb|\('(.*)\(efsa', 'foodex2\)'\)|
    \end{enumerate}
    followed by removal of leading and trailing whitespaces. 
    Note that concepts in this ontology sometimes have an empty literal given by \texttt{rdf:label}; in these cases, the annotation properties \texttt{obo:hasSynonym} and \texttt{obo:hasExactSynonym} are used instead.
    \item \go: no individual processing.
\end{itemize}


\section{Object Property Verbalisation}
\label{appendix:object-prop}

Different from verbalising an atomic concept where we simply use its name (or alias), we enforce some simple rules to verbalise an object property for a basic grammar fix. If the property name starts with a \textit{passive verb}, \textit{adjective}, or \textit{noun}, we append \state{is} to the head. For example, \state{characteristic of} is changed to \state{is characteristic of}; \state{realised in} is changed to \state{is realised in}. Note that the word's part-of-speech tag is automatically determined using the Python library \texttt{Spacy}\footnote{\url{https://spacy.io/}}.


\section{Complex Concept Verbalisation Examples} \label{appendix:data-examples}

For clearer understanding of how our verbalisation approach works,
we present some typical examples of verbalised concepts from the constructed Complex SI datasets in Table \ref{tab:data-examples}.

\section{Implementation Choices for Assumed Disjointness} \label{appendix:assumed-disjointess-alternative}

As mentioned in Section~\ref{sec:SI-general}, validating the disjointness axiom for each concept pair $(C, D)$ we have sampled as a potential negative subsumption would be time-consuming because we need to iteratively add the disjointness axiom into the ontology $\mathcal{O}$, conduct reasoning, and remove the axiom afterwards. Therefore, in practice we can use the following conditions to replace the satisfiability~check:
\begin{enumerate}[\itshape(i)]
    \item \textbf{No subsumption relationship}: $\mathcal{O} \not\models C \sqsubseteq D$ and $\mathcal{O} \not\models D \sqsubseteq C$;
    \item \textbf{No common instance}: there is no \textit{named} instance $a$ in $\mathcal{O}$ such that $\mathcal{O} \models C(a)$ and $\mathcal{O} \models D(a)$.
\end{enumerate}
If $C$ and $D$ satisfy these two conditions, they are \textbf{likely to be satisfiable} after adding the disjointness axiom $C \sqcap D \sqsubseteq \bot$ into $\mathcal{O}$. Since these two conditions involve \textbf{no extra reasoning} for a new axiom, they are much more efficient than iteratively conducting satisfiability check for candidates. 

It is important to notice that we still need the \textbf{no common descendant} check to prevent foreseeable unsatisfiability. 

\begin{enumerate}
[\itshape(i)]
\addtocounter{enumi}{2}
    \item \textbf{No common descendant}: there is no named atomic concept $A$ in $\mathcal{O}$ such that $\mathcal{O} \models A \sqsubseteq C$ and $\mathcal{O} \models A \sqsubseteq D$.
\end{enumerate}

\noindent This is because if
there is a named atomic concept $A$ that is an inferred sub-class (i.e., descendant) of $C$ and $D$, then it is possible that $C$ and $D$ are satisfiable in $\mathcal{O} \cup \{ C \sqcap D \sqsubseteq \bot \}$, but $A$ is certainly unsatisfiable (equivalent to $\bot$).

\begin{figure}[t!]
\begin{center}
    \includegraphics[width=0.3\textwidth]{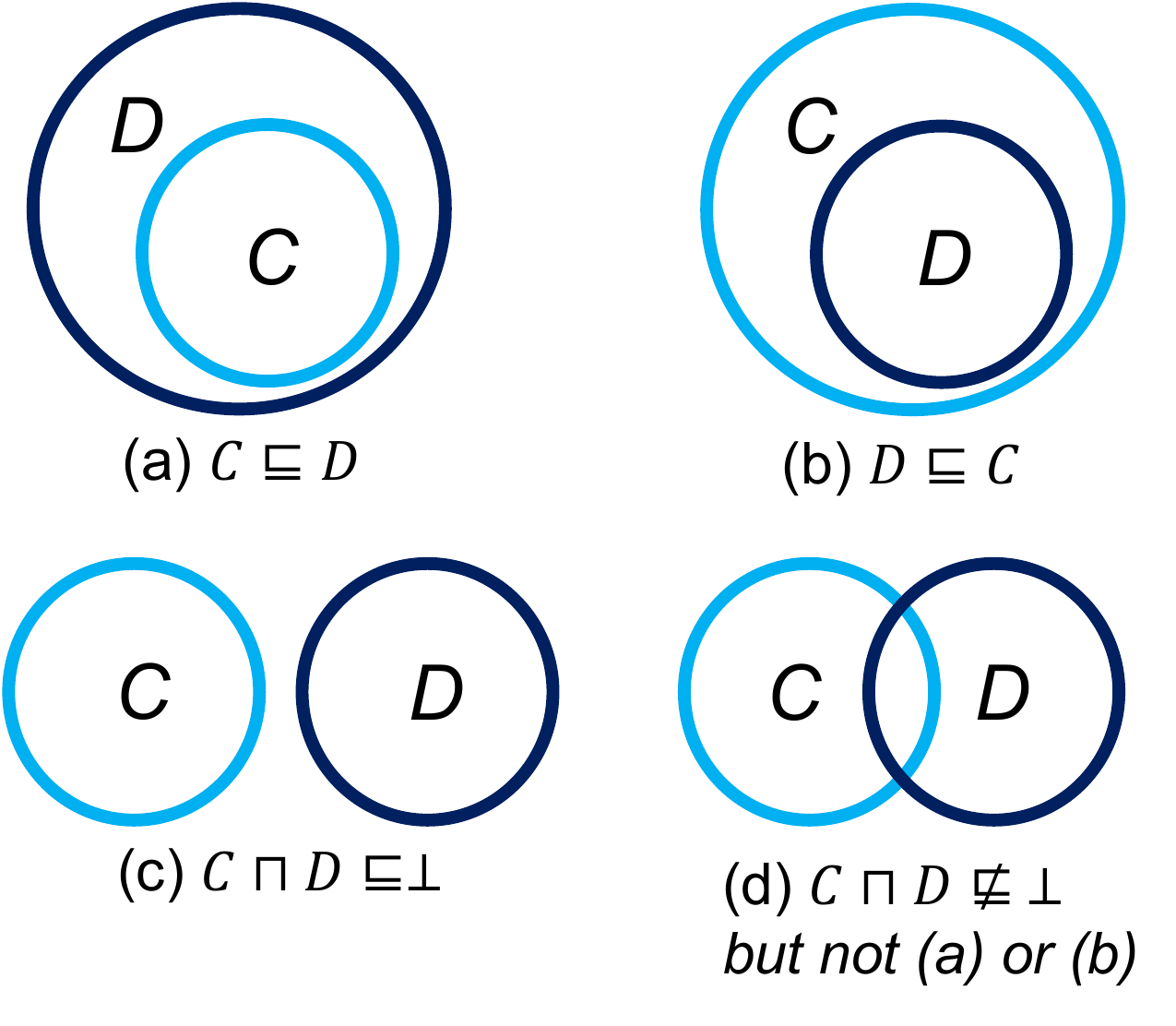}
\caption{Set-based semantics for relationships between two ontology concepts.}
\label{fig:explain-subs}
\end{center}
\end{figure}

\begin{table*}[ht!]
    \centering
    \renewcommand{\arraystretch}{1.2} 
    {\fontsize{9}{10.8}\selectfont
    \begin{tabularx}{0.9\textwidth}{X Y Y Y Y Y Y Y}
    \toprule
     & & \multicolumn{4}{c}{\textbf{Atomic SI}} & \multicolumn{2}{c}{\textbf{Complex SI}}\\
    \cmidrule(lr){3-6} \cmidrule(lr){7-8}
    \textbf{Setting} & \textbf{biMNLI} &  \textbf{Schema.org} & \textbf{DOID} & \textbf{FoodOn} & \textbf{GO} & \textbf{FoodOn} & \textbf{GO}\\
    \midrule
    \multicolumn{2}{l}{\textsf{roberta-base}} \\
    \hspace{3mm} K=0   &  62.5 (6.5) &  56.4 (3.6) &  53.3 (4.0) &  54.6 (4.4) &  49.0 (2.4) &    55.9 (3.6) &  48.7 (3.1) \\
    \hspace{3mm} K=4   &  67.6 (5.2) &  62.9 (5.2) &  61.8 (6.7) &  62.1 (4.2) &  65.2 (5.0) &    62.4 (3.2) &  52.2 (7.1) \\
    \hspace{3mm} K=8   &  70.7 (4.5) &  71.2 (4.5) &  72.9 (5.7) &  69.0 (5.2) &  70.4 (5.1) &    66.0 (4.4) &  63.0 (5.0) \\
    \hspace{3mm} K=16  &  74.3 (3.3) &  79.7 (4.2) &  83.4 (2.5) &  79.8 (3.0) &  78.3 (3.0) &    70.2 (5.5) &  73.8 (4.0) \\
    \hspace{3mm} K=32  &  78.8 (1.1) &  84.3 (2.0) &  89.0 (1.4) &  85.0 (1.1) &  84.6 (2.5) &    77.0 (1.5) &  76.4 (2.5) \\
    \hspace{3mm} K=64  &  80.9 (1.5) &  88.3 (1.5) &  91.2 (0.7) &  88.2 (0.7) &  87.3 (0.8) &    80.0 (2.0) &  81.7 (1.4) \\
    \hspace{3mm} K=128 &  85.1 (1.0) &  91.1 (0.7) &  92.4 (0.7) &  90.0 (0.7) &  89.0 (0.8) &    85.5 (1.3) &  86.9 (1.5) \\
    \midrule
    
\multicolumn{2}{l}{\textsf{roberta-large}} \\
    \hspace{3mm} K=0   &  68.7 (6.2) &  61.7 (7.2) &  59.8 (5.4) &  60.1 (8.8) &  54.6 (1.9) &    56.1 (1.9) &  50.4 (0.6) \\
    \hspace{3mm} K=4   &  78.1 (6.6) &  69.4 (5.4) &  74.0 (5.5) &  71.6 (4.4) &  67.6 (3.4) &    64.1 (5.1) &  56.9 (5.7) \\
    \hspace{3mm} K=8   &  83.0 (5.2) &  78.5 (3.0) &  84.4 (3.8) &  77.0 (6.0) &  75.3 (3.2) &    71.3 (3.1) &  64.2 (7.6) \\
    \hspace{3mm} K=16  &  87.5 (2.4) &  84.4 (2.4) &  87.6 (2.3) &  83.4 (3.5) &  82.8 (1.9) &    76.2 (2.5) &  76.4 (3.3) \\
    \hspace{3mm} K=32  &  89.9 (1.2) &  87.3 (1.9) &  92.3 (0.7) &  88.9 (1.6) &  87.7 (1.6) &    80.8 (3.8) &  81.6 (2.2) \\
    \hspace{3mm} K=64  &  90.8 (1.4) &  90.4 (0.8) &  92.6 (0.7) &  90.9 (1.2) &  90.1 (0.7) &    84.1 (1.4) &  86.2 (2.2) \\
    \hspace{3mm} K=128 &  93.0 (0.8) &  92.9 (0.8) &  93.4 (0.5) &  92.2 (0.5) &  91.0 (0.7) &    88.4 (1.1) &  90.2 (1.0) \\
    \cmidrule(lr){1-8}
    \hspace{3mm} full  &  97.5 &  95.4  &  97.8  &  98.7  &  98.1  &    95.8  &  98.8  \\
    \bottomrule
    \end{tabularx}}
    \caption{\label{tab:full-roberta} \footnotesize Full results of \textsf{roberta-base} and \textsf{roberta-large} on the biMNLI, Atomic SI, and Complex SI tasks with each cell stating \textit{``mean accuracy (standard deviation)''} except for the majority vote and fully supervised settings where standard deviation is not available.}
    \vspace{-4mm}
\end{table*}

\begin{figure*}[ht!]
\begin{center}
\includegraphics[width=0.82\textwidth]{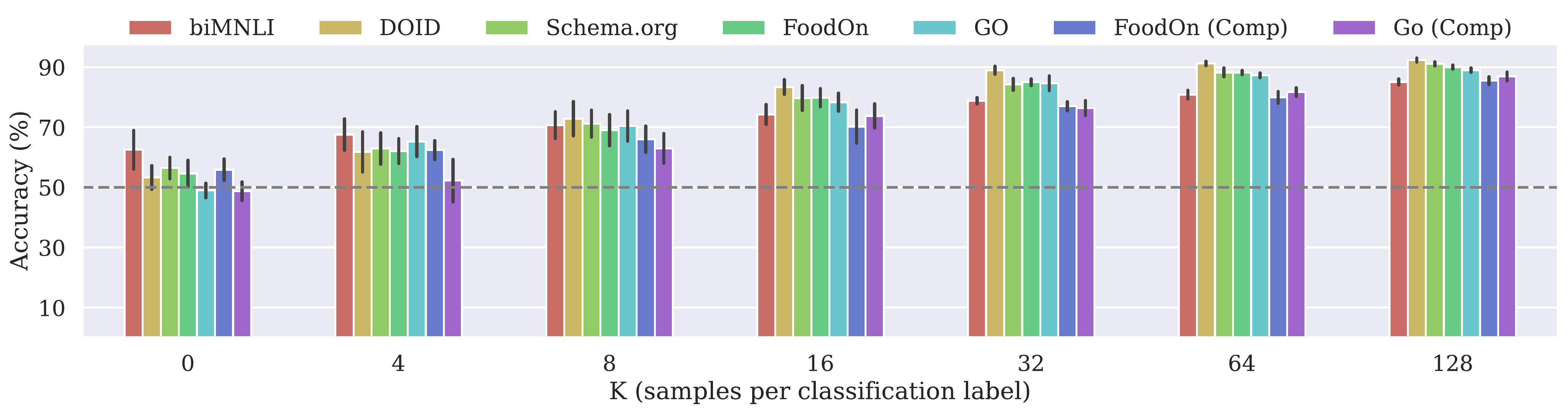}
\caption{\footnotesize Visualisation of K-shot results (for \textsf{roberta-base}) on the biMNLI, Atomic SI, and Complex SI tasks where the dotted horizontal line indicates majority vote. The order of the bars is the same as in the legend.}
\label{fig:roberta-base}
\end{center}
\vspace{-4mm}
\end{figure*}

\section{Set-based Interpretations of Subsumption Samples} \label{appendix:explain-subs}

In this section, we provide more explanation for how we define positive and negative samples in the Subsumption Inference (SI) task.

Recall the definitions in Section \ref{sec:onto}, an ontology $\mathcal{O}$ entails a subsumption axiom $C \sqsubseteq D$ if it holds for \textbf{every interpretation} $\mathcal{I}$ of $\mathcal{O}$. In terms of set-based semantics, this refers to case \textit{(a)} in Figure \ref{fig:explain-subs}. 
In the \textit{(b)}, \textit{(c)}, or \textit{(d)} cases, there exists at least one interpretation $\mathcal{I}$, such that we can find an individual $x$ that $\inter{x} \in \inter{C}$ and $\inter{x} \not\in \inter{D}$;
hence $\mathcal{O}$ \textbf{does not entail} the subsumption axiom $C \sqsubseteq D$. Non-subsumption is entailed only when \textit{(a)} \textbf{does not hold for every interpretation} of $\mathcal{O}$.

Disjointness corresponds to \textit{(c)} in Figure \ref{fig:explain-subs} where the set of $C$ and the set of $D$ have no overlap for every interpretation. Non-subsumptions an ontology \textbf{typically} entails come from the disjointness axioms (but disjointness $\forall x. C(x) \rightarrow \neg D(x)$ is \textbf{stricter} than non-subsumption $\exists x. C(x) \land \neg D(x)$). Nevertheless, ontologies are typically underspecified in terms of disjointness, and thus getting enough negative samples is unfeasible.
To find a middle ground, it is reasonable to adopt heuristics.
The assumed disjointness we follow in Section \ref{sec:SI-general} in the main body of the paper serves this purpose. In the ideal setting where we check the \textbf{satisfiability} of $C$ and $D$ after adding the disjointness axiom and \textbf{no common descendant} of $C$ and $D$, cases \textit{(a)} and \textit{(b)} in Figure \ref{fig:explain-subs} will be prevented and the chance of \textit{(d)} reduced. Even in the practical alternative proposed in this Appendix~\ref{appendix:assumed-disjointess-alternative}, the \textbf{no subsumption relationship} condition also ensures that \textit{(a)} and \textit{(b)} are not entailed and the \textbf{no common descendant} and \textbf{no common instance} conditions reduce the chance of~\textit{(d)}. Thus, the assumed disjointness is a reasonable approach to approximate non-subsumptions.

\section{Complementary Results and Figures} \label{appendix:complement}
In the main body of the paper, we report partial results (accuracy scores and standard deviations) of \textsf{roberta-large} and \textsf{roberta-base} for $K \in \{0, 4, 32, 128\}$. In Table \ref{tab:full-roberta}, we present full results of both LMs for $K \in \{0, 4, 8, 16, 32, 64, 128\}$.

Besides, we provide the visualisation of $K$-shot results for \textsf{roberta-base} in Figure \ref{fig:roberta-base}.
The observations are consistent with those for \textsf{roberta-large} in Figure \ref{fig:roberta-large}.

\end{document}